\documentclass[smallcondensed, final]{svjour3}     
\smartqed  
\usepackage{graphicx}
\usepackage{amsmath}
\usepackage[utf8]{inputenc} 
\usepackage[T1]{fontenc}
\usepackage{amssymb}
\usepackage{booktabs}
\usepackage{dsfont}
\usepackage{mathtools}
\usepackage{nameref}
\usepackage[numbers, sort]{natbib}

\journalname{SN Computer Science}

\begin{document}

\title{Implications on Feature Detection when using the Benefit-Cost Ratio}


\author{Rudolf Jagdhuber \and Jörg Rahnenführer*}


\institute{Rudolf Jagdhuber \at
              Department of Statistics, TU Dortmund \\
              Vogelpothsweg 87\\
              44227 Dortmund\\
              ORCID: 0000-0002-2958-2716
           \and
           Jörg Rahnenführer* (Corresponding Author) \at
              Department of Statistics, TU Dortmund \\
              Vogelpothsweg 87\\
              44227 Dortmund\\
              \email{rahnenfuehrer@statistik.tu-dortmund.de}
}

\date{Received: date / Accepted: date}

\maketitle

\begin{abstract}
In many practical machine learning applications, there are two objectives: one 
is to maximize predictive accuracy and the other is to minimize costs of the 
resulting model. These costs of individual features may be financial costs, but 
can also refer to other aspects, like for example evaluation time. Feature 
selection addresses both objectives, as it reduces the number of features and 
can improve the generalization ability of the model. If costs differ between 
features, the feature selection needs to trade-off the individual benefit and 
cost of each feature. A popular trade-off choice is the ratio of both, the 
BCR~(benefit-cost ratio). In this paper we analyze implications of using this 
measure with special focus to the ability to distinguish relevant features from 
noise. We perform a simulation study for different cost and data settings and 
obtain detection rates of relevant features and empirical distributions of the 
trade-off ratio. Our simulation study exposed a clear impact of the cost 
setting on the detection rate. In situations with large cost differences and 
small effect sizes, the BCR missed relevant features and preferred cheap noise 
features. We conclude that a trade-off between predictive performance and costs 
without a controlling hyperparameter can easily overemphasize very cheap noise 
features. While the simple benefit-cost ratio offers an easy solution to 
incorporate costs, it is important to be aware of its risks. Avoiding costs 
close to 0, rescaling large cost differences, or using a hyperparameter 
trade-off are ways to counteract the adverse effects exposed in this paper.

\end{abstract}

\section{Background}\label{sec:int}
Feature selection is a common preprocessing step in many learning tasks, which 
aims to remove noise features and to identify a suitable subset of relevant 
information from an often high dimensional data set. This way it can improve 
the generalization ability and reduces computational complexity of subsequent 
learning algorithms. Cost-sensitive learning describes an extension of this 
general feature selection problem by introducing acquisition costs for selected 
features. Depending on the application field, these costs may not only refer to 
financial aspects, but could also represent a time span to raise a feature or a 
patient harm during the sample taking process.

The general strategy to incorporate feature costs into a feature selection 
framework depends on the problem at hand. If a fixed total feature cost limit 
can be defined, the problem reduces to an additional optimization constraint 
for the feature selection problem. Many example applications of fixed budget 
costs can be found \citep{min2011test, min2014feature, min2016semi, 
liu2011genetic, jagdhuber2020cost}.
For situations without a fixed budget, the goal may be to harmonize costs of 
features and costs of prediction errors by identifying an optimal trade-off. 
Research on these flexible solutions can be found, e.g., in \citet{zhou2016cost} 
or \citet{bolon2014framework}. A third situation is given when feature 
acquisition is undertaken sequentially. In such situations, tests can take 
advantage of intermediate results and reduce total costs by only requesting 
further features if the benefit justifies the additional cost, see, e.g.
\citep{xu2014classifier, xu2013cost, kusner2014feature}.

A common factor for all mentioned tasks is the need to somehow trade-off 
the benefit of a feature with its cost. As these measures are on different 
scales, the main options to combine them are either to optimize the ratio of 
both~\citep{min2011test,min2014feature,min2016semi,leskovec2007cost,
grubb2012speedboost,paclik2002feature}, or to trade-off a weighted sum 
\citep{bolon2014framework, xu2014classifier, xu2013cost, kusner2014feature,
kong2016beyond, xu2012greedy}.
In this paper we take a closer look at the first mentioned alternative. We 
analyze the consequences of using a simple benefit-cost ratio (BCR) with 
respect to the detection of relevant features against noise. 

We start by defining the general cost-sensitive feature selection problem
and discussing the theoretical implications of using the BCR. In the following
section we perform a simulation study to analyze the influence of multiple
data parameters and feature cost settings on the feature detection rate. 
Finally, we present the obtained results, discuss the general applicability of 
the basic BCR and provide recommendations for alternative trade-off measures.

\section{Problem Definition}\label{sec:met}
Given is a data set with $n$ observations $D_i, i = 1,\dots, n$ and $p$ 
features $x_{ij}, j = 1,\dots, p$ for observation $i$, and continuous response 
$y_i$ for observation $i$. Assume that the true relation is given by 
$y_i = \beta_0 + \sum_{j = 1}^{p_\text{rel}}x_{ij}\beta_j + \epsilon_i$, with 
$\epsilon_i\sim \mathcal{N}(0, \sigma^2)$. In this data $p_\text{rel}$ features
are assumed to have an influence on the response, while all other 
$p_\text{noise}$ features are independent of $y$. Then the goal of feature 
selection is to identify the subset of relevant features.

One obvious approach to ensure finding this optimal subset is an exhaustive 
search, i.e. to consider all possible subsets. However, this approach is 
usually not feasible for high dimensional feature spaces. Thus, heuristic 
selection algorithms like greedy sequential forward selection (SFS) are used. 
SFS iteratively adds the single most promising feature to the current result 
set. A typical way to estimate the importance of a feature $x_j$ when added to 
a given set $s$ is to calculate the performance gain of a statistical model 
including the feature $M(s\cup\{x_j\}|D)$ compared to a baseline model without 
it $M(s|D)$. The feature with the highest gain in performance is then selected. 
Assuming a performance criterion $Q$, for which the optimal value is the 
minimal value, we can formulate one feature selection step of SFS by
\begin{equation}\label{eq:fs}
    \hat{m} = 
    \underset{j}{\arg\max}\left\{Q(M(s|D)) - Q(M(s\cup\{x_j\}|D))\right\} 
    \eqqcolon\underset{j}{\arg\max}~\Delta Q_j .
\end{equation}
In many real-world scenarios, obtaining a feature $x_j$ may cause individual 
feature costs $c_j$. Cost-sensitive feature selection aims to incorporate these 
costs into the selection process to find cheap \textit{and} well performing 
models. A popular method is to adapt the problem of Equation~(\ref{eq:fs}) to
\begin{equation}\label{eq:cfs}
    \hat{m} = \underset{j}{\arg\max}~\frac{\Delta Q_j}{c_j}.
\end{equation}
This ratio of benefit and cost leads to a simple trade-off optimization, which
relates the importance of a feature to its cost. In the following we describe  
negative implications of this simple and popular method when discriminating 
between relevant and noise features.

The true performance gain of a \emph{noise} feature is a value smaller or equal 
to zero, as it has no relation to the response but may create additional 
uncertainty. The true performance gain of a \emph{relevant} feature is 
typically a value greater than 0. Nevertheless, the actual performance gain 
estimated on a sample data set does not always result in these true values. It 
can rather be seen as a random variable following a certain unknown 
distribution around the true value:
\begin{equation}
    \Delta Q_j \sim \mathcal{V}_j(\cdot)    
\end{equation}
For a real world data situation, the theoretical distributions of $\Delta Q_j$ 
for different $j$ can be assumed to overlap to some extent. That means, for one 
given sample data set, the actual estimated performance gain of a noise feature 
may be higher than the one of the relevant feature and thus an irrelevant 
feature may be selected.

When incorporating cost according to Equation~(\ref{eq:cfs}), the performance 
gain distribution of feature $x_j$ is scaled by a positive factor $c_j$, which 
increases and broadens $\mathcal{V}_j$, if $c_j < 1$, and decreases and narrows 
it, if $c_j > 1$. Increasing and broadening the distribution of a noise 
feature, while not altering the one of a relevant feature increases the overlap 
of both distributions. Therefore the probability of falsely selecting the noise 
feature increases. In some situations this problem may be negligible. In 
others, the cost-sensitive feature selection procedure can completely obfuscate 
any relevant feature.

The actual magnitude of the cost influence depends on many factors including 
the sample size $n$, the true effect size of relevant features $\beta$, the 
residual variance $\sigma^2$, the statistical model, and the performance 
measure $Q$. The goal of this paper is to analyze this problem and describe 
multiple parameter settings and their influence on the feature detection rate. 
We focus on linear regression models and use the root mean squared error (RMSE) 
on independent data to assess the quality of models. The RMSE is defined as
\begin{equation}\label{eq:AIC}
\text{RMSE} = \sqrt{\sum_{i = 1}^n\left(y_i - \hat{\beta}_0 - 
\sum_{j\in s}x_{ij}\hat{\beta}_j\right)^2},
\end{equation}
with $\hat{\beta}_0$ and $\hat{\beta}_j$ estimated on training data and 
$x_{ij}$ and $y_i$ denoting observations of an independent test data set. By 
using such an independent test data set, the RMSE also allows a result of no 
improvement after adding a feature.

In the following, for ease of presentation, we describe a single feature 
selection step of SFS from a pool of $p_\text{rel}$ relevant and 
$p_\text{noise}$ noise features. We also define this single step to be the 
first selection step, i.e. we define our baseline model to be the intercept 
model and compare the quality of all one-feature models. The final selection 
result of this one step can either be 'noise selected', 
'relevant feature selected', or 'no feature selected'. Similarly to 
Definition~(\ref{eq:fs}), in the following we denote the gain in RMSE for 
feature j by $\Delta\text{RMSE}_j$. The corresponding distribution 
$\mathcal{V}_j(\cdot)$ has no analytical form. In the following simulation 
study, we overcome this problem by numerically approximating this distribution 
and computing selection probabilities on the empirical distribution.

\section{Simulation Study}\label{sec:sim}
The goal of our simulation study is to assess the detection rate of a 
cost-sensitive feature selection step in multiple parameter settings. 
Additionally we aim to analyze the empirical distribution of our performance 
measure to further illustrate effects of cost scaling. We consider a linear 
regression scenario. Our response variable, as well as all $p$ features are 
assumed to be normally distributed. We define $p_\text{rel}$ features to be 
relevant and the remaining $p_\text{noise} = p - p_\text{rel}$ features to be 
noise. The individual costs of features can be seen as a relative scaling 
between the respective $\Delta\text{RMSE}_j$ values of the features. To 
simplify our analyses, we do not consider individual costs for all features, 
but define only one single scaling factor $\theta$ for the relevant features. 
Hence, we implicitly define equal costs for the group of noise features and 
equal costs for the group of relevant features. We only differentiate between 
costs for information and costs for noise. To thoroughly assess the influence 
on the detection rate, we vary the feature cost scaling factor $\theta$ between 
1, 10, 100 and 1000, the number of relevant features $p_\text{rel}$ between 1, 
2, 5 and 10, the number of noise features $p_\text{noise}$ between 1, 10 and 
50, and the effect size of the relevant feature $\beta$ between 
$0, 0.01, \dots, 0.5$. For multiple relevant features, we do not vary the 
effect size and define $\beta_j\coloneqq\beta$. 

For each parameter combination, $B = 1000$ training ($n_\text{train} = 100$) 
and test data sets ($n_\text{test} = 1000$) are generated as follows. In a 
first step, features are drawn from a $p$-dimensional normal distribution
\begin{equation}\label{eq:drawX}
x_1, \dots, x_p \sim \mathcal{N}_p(\mathbf{0}, \mathbf{I}_p),
\end{equation}
where $\mathbf{I}_p$ is the $p$-dimensional identity matrix. Next, the response 
is drawn from the normal distribution
\begin{equation}\label{eq:drawY}
y_i\sim \mathcal{N}\!\left(\beta_0 + \sum_{j = 1}^{p_\text{rel}}x_{ij}\beta~,~ 
\sigma^2\right).
\end{equation}
We set the intercept to $\beta_0 = 1$ and the residual variance to 
$\sigma^2 = 1$ for all settings.

For every data set obtained in this way, we fit the baseline intercept model 
and all one-feature models separately and obtain
\begin{equation} 
\begin{split}
M_0&: y = \hat\beta_0 + \epsilon, \\ 
M_j&: y = \hat\beta_0 + x_j\hat\beta + \epsilon,\;\; j = 1,\dots, p. 
\end{split}
\end{equation}
We then compute the increase in RMSE for all features by 
\begin{equation}
\Delta\text{RMSE}_j = \text{RMSE}(M_0) - \text{RMSE}(M_j), \;\; j = 1,\dots, p.
\end{equation}
As we are only interested in the question if a noise feature or a relevant 
feature is selected, we define the RMSE gain of noise and relevant features as 
our target variables. The best $\Delta\text{RMSE}$ value indicates the 
candidate that is selected from the noise and the relevant features, 
respectively.
\begin{equation}\label{eq:empdist}
  \begin{split}
  &\Delta\text{RMSE}_\text{rel} = \max\left(\{\Delta\text{RMSE}_j : 
  j = 1,\dots,p_\text{rel}\}\right)\\
  &\Delta\text{RMSE}_\text{noise} = \max\left(\{\Delta\text{RMSE}_k : 
  k = p_\text{rel} + 1,\dots, p\}\right)
  \end{split}
\end{equation}
As described earlier, we define our cost setting by a single factor $\theta$, 
which scales relevant features. Hence, the assessed measure of RMSE gain for 
relevant features actually results in 
$\frac{\Delta\text{RMSE}_\text{rel}}{\theta}$.

The final feature selection on a single data set can lead to three different 
outcomes $\hat{m}$. We only consider increases in $\Delta\text{RMSE}$. 
Therefore, if neither relevant, nor noise features result in a positive RMSE 
gain, then no feature is selected.
\begin{equation}\label{eq:fit}
  \hat{m} = \arg\max\left(\frac{\Delta\text{RMSE}_\text{rel}}{\theta}, 
  \Delta \text{RMSE}_\text{noise}, 0\right)
\end{equation}
As every setting is repeated 1000 times with newly simulated data sets, we can 
estimate the probability for each selection result $m$ by looking at the 
relative frequency among those 1000 runs. We can further obtain empirical 
distributions of $\Delta\text{RMSE}$ for relevant and for noise features in 
different settings. The results for both of these analyses are presented in the 
following section.

\section{Results}\label{subsec:simResults}
\begin{figure*}\centering
\includegraphics[width = \textwidth]{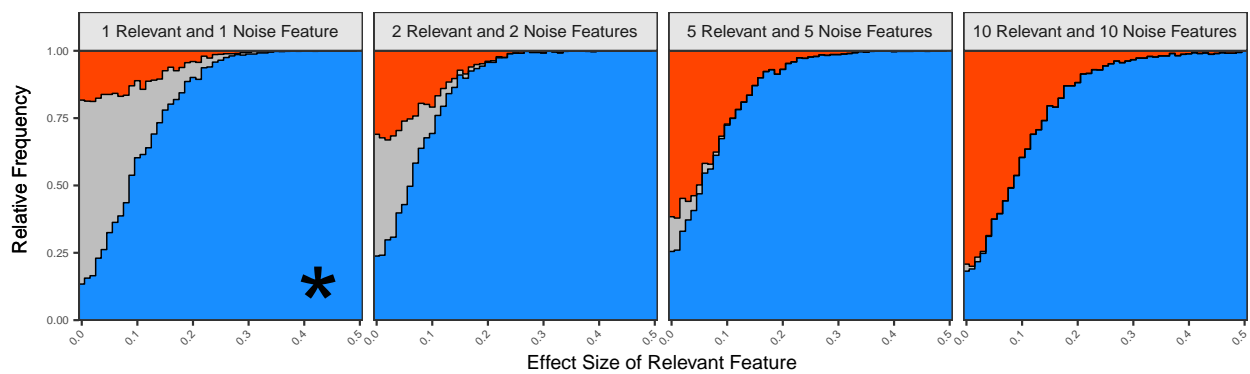}
\caption{Selection probabilities of relevant, noise, or no feature, along 
multiple values of the cost-scaling factor $\theta$ (columns), the number of 
noise features $p_\text{noise}$ (rows) and the effect size of relevant features 
$\beta$ (x-axis per plot). The main 3x4 plot matrix analyzes the setting of 
$p_\text{rel} = 1$. An additional bottom row illustrates corresponding plots 
for different numbers of relevant features $p_\text{rel}$ at a fixed scaling 
level $\theta=10$. The plots annotated with a star are 
identical.\label{fig:selprob}}
\end{figure*}

This section comprises the analysis of the selection probabilities with main 
results presented in Figure~\ref{fig:selprob} and the analysis of the empirical 
distribution of the selection criterion with main results presented in 
Figure~\ref{fig:empdist}. 
To provide comprehensive illustrations, both analyses focus mainly on the 
setting with one relevant feature, and only a small analysis to describe the 
effects of different numbers of relevant features is added. Corresponding 
illustrations of all settings can be found in the Appendix.

\subsection{Feature Detection Rates}
The individual plots of Figure~\ref{fig:selprob} illustrate the estimated 
probabilities for the three selection outcomes 
'\textit{relevant feature selected}', '\textit{noise feature selected}' and 
'\textit{no feature selected}' along multiple effect sizes of the true effect 
$\beta$. Rows of the main plot matrix relate to different numbers of noise 
features, while columns represent the extent of cost-scaling applied to the 
relevant feature. 

The top-left plot describes a setting with one relevant and one noise feature.
No cost scaling is applied, which could refer to a setting without or with 
equal costs, respectively. At an effect size of $\beta = 0$, where both 
features can be considered noise, their selection probability is approximately 
equal. In almost 70\% of the cases, neither of them is selected. When 
increasing the effect size $\beta$, the selection probability for the relevant 
feature rises, while the probabilities for both other outcomes decrease. From 
around $\beta = 0.3$ onward, the relevant feature is identified approximately 
100\% of times.

Increasing the number of noise features (rows 2 and 3) changes this result in 
multiple ways. The main difference can be seen in the number of times that no 
feature is selected. This value is reduced greatly for 10 noise features and 
disappears completely for 50 noise features. The other difference is that the 
selection curve of the relevant feature starts at a lower value and reaches 
100\% selection slightly later. These differences are however more subtle.

The main focus of our paper lies on the effect of incorporating costs and thus
scaling the performance distribution of the relevant features. This scaling 
factor corresponds to the columns of the plot matrix. When increasing the  
factor, the decrease in selection probability of noise for higher effect sizes 
becomes smaller, eventually resulting in an approximately constant noise 
selection probability at $\theta = 1000$. As the initial selection probability 
for noise increases with a larger number of noise features, the combined effect 
results in always selecting noise at the bottom-right plot.

The effects of increasing the number of relevant features is illustrated for a
fixed scaling factor $\theta = 10$ and an equal number of noise and relevant 
features in the additional bottom row of Figure~\ref{fig:selprob}. The main 
observation is that the extent of selecting no feature reduces with increasing 
$p_\text{rel}$ and instead a noise feature is selected. The probability of 
selecting a relevant feature does not seem to be strongly influenced, it is 
only slightly pushed back by noise and reaches the area of 100\% selection for 
slightly larger efect sizes. Full illustrations including multiple values of 
$\theta$ and non-identical $p_\text{rel}$ and $p_\text{noise}$ are given 
in \nameref{add:1}.

\subsection{Empirical Distribution of $\Delta\text{RMSE}$}
\begin{figure*}
 \includegraphics[width = \textwidth]{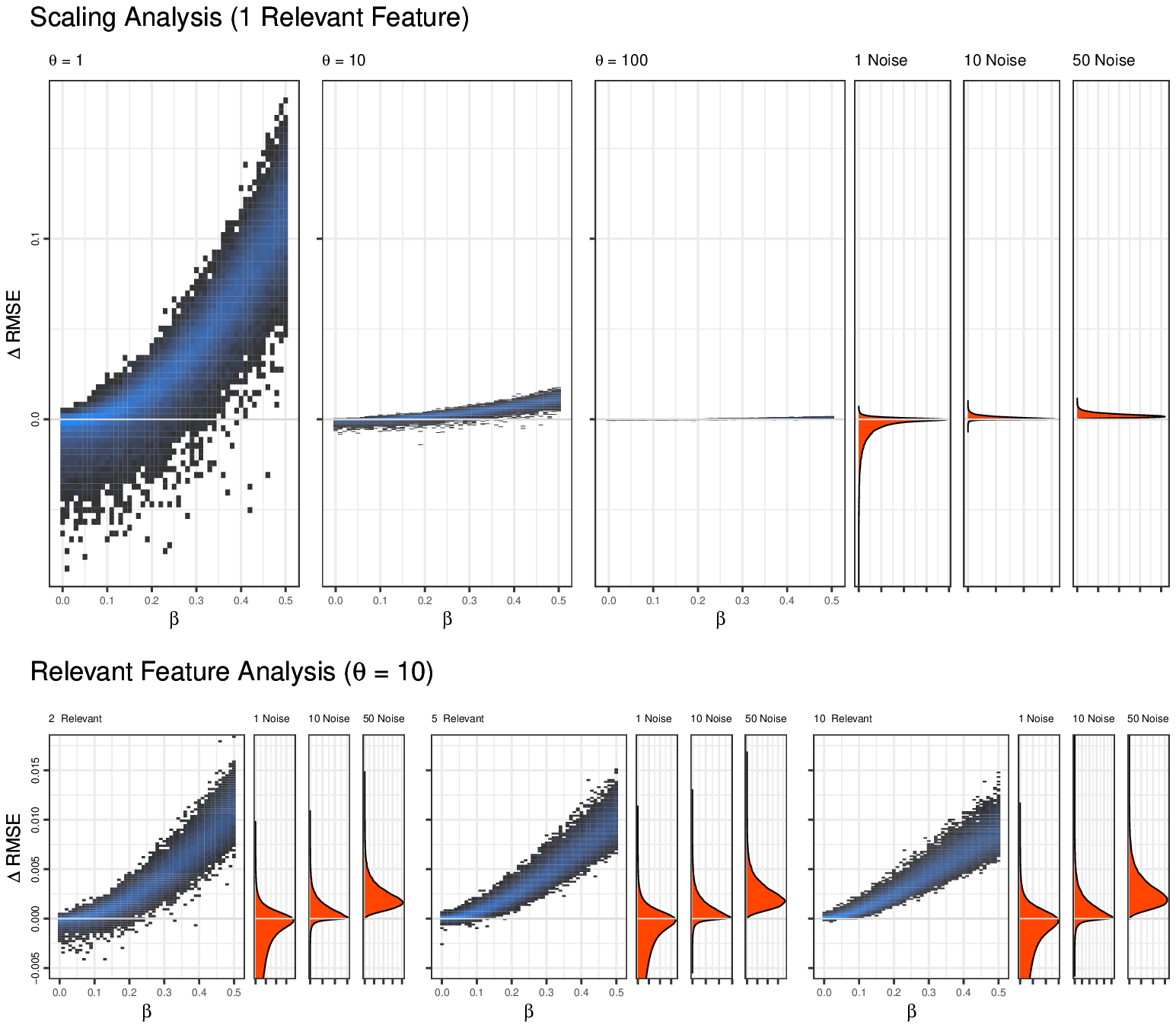}
  \caption{Empirical distributions of $\Delta\text{RMSE}$ for noise features 
  (red) and relevant features (blue). The latter are illustrated as heatmap 
  along different values of the true effect size $\beta$. Lighter colors 
  indicate higher densities. The first row describes a setting with 
  $p_\text{rel} = 1$. Three plots of relevant features for different values of 
  the cost-scaling $\theta$ and three plots of noise features for different 
  values of $p_\text{noise}$ are given. The bottom row shows corresponding 
  plots for different values of $p_\text{rel}$ at a fixed cost-scaling of 
  $\theta = 10$.\label{fig:empdist}}
\end{figure*}

The empirical distribution of RMSE gain for the relevant features depends on the
true effect $\beta$, the cost scaling parameter $\theta$, and the number of 
relevant features in the model. For noise, it only depends on the numbers of 
noise and relevant features, as the true effect is 0 and no scaling of noise is 
performed. A comprehensive illustration of all analyzed distributions for 
$p_\text{rel} = 1$ is given in the top plot of Figure~\ref{fig:empdist}. A 
heatmap describes the distributions of RMSE gain for relevant features along 
different effect sizes. Lighter colors correspond to higher densities. RMSE 
gains for noise features are depicted by three density curves for settings with 
1, 10 and 50 noise features, respectively. A gray area highlights the decision 
boundary for not selecting any feature. 

The given plots provide deeper insight into the selection decisions illustrated 
previously in Figure~\ref{fig:selprob}. Analyzing the noise features, the 
distribution of RMSE gains of one single noise feature has the great majority 
of its probability mass within the gray area and would not be selected, 
regardless of the RMSE gain of the relevant feature. However, when increasing 
the number of noise fatures $p_\text{noise}$, the noise distribution steadily 
moves out of this area. For the relevant parameter, the unscaled distribution 
(top-left plot) increases superlinearly along $\beta$ and completely passes any 
noise distribution at around $\beta=0.4$. A cost-scaling however lowers the 
slope of this increase and decreases the variance of the relevant feature 
distribution. As a consequence of both, surpassing the noise distributions 
happens notably slower. For $\theta = 100$, the size of the relevant feature 
distribution compared to noise is shrunken down to a level making it almost 
invisible in the plot. The largest noise distribution is not surpassed at all 
in our range of $\beta$ values. However, an important observation is that the 
total density of $\Delta\text{RMSE}_\text{rel}$ below or equal to zero is 
constant for any scaling. We omitted an illustration for $\theta = 1000$ as it 
is invisible on this scale. Rescaled versions for all distributions can be 
found in \nameref{add:2}.

The bottom part of Figure~\ref{fig:empdist} depicts the effects of increasing 
the number of relevant features in the true model, for $\theta = 10$. Mainly, 
the general density mass below zero decreases when the number of relevant 
features increases. However, the maximum $\Delta\text{RMSE}$ value for 
$\beta = 0.5$ also decreases. The RMSE gain of noise features on the other hand 
results in almost identical density curves.

\section*{Discussion and Conclusion}\label{sec:dis}
The simulation study revealed multiple consequences of cost-sensitive feature 
selection when using the popular benefit-cost ratio without a hyperparameter. 
In Figure~\ref{fig:selprob}, we see that cost-scaling the benefit 
$\Delta\text{AIC}$ makes the selection probability of noise features more 
robust, especially for large true effects. With $\theta \to \infty$, this 
probability becomes independent of $\beta$. However, the frequency of selecting 
noise does not necessarily approach 1, but converges to a certain limit. For 
$\theta \to \infty$, this limit is given by 
$P(\Delta\text{RMSE}_\text{noise} > 0)$. Values with negative RMSE difference 
will never be selected, regardless of the scaling. With an increasing number of 
noise features, the probability that all estimated performance gains are 
negative decreases. Hence, the described limit for selecting noise rises. The 
third row of Figure~\ref{fig:selprob} illustrates the consequences of both 
effects, which eventually results in a noise selection probability of 
approximately 1 for all $\beta$ values. The empirical distributions shown in 
Figure~\ref{fig:empdist} further describe this relation. With higher cost 
penalization, the slope and variance of the RMSE gain distribution along 
$\beta$ decreases. The probability regions favoring noise over the relevant 
feature constantly become larger as $\theta$ increases, yet the probability 
masses above and below 0 stay constant, further illustrating the probability 
limit of noise selection. The effects of increasing the number of relevant 
features in the true model are more subtle. The selection probability plots 
mainly show the effects already observed when increasing the number of noise 
features. The differences in the empirical densities of RMSE gains of relevant 
features in Figure~\ref{fig:empdist} are the result of two effects. On the one 
hand, the maximum RMSE results in a higher value for a higher number of 
features. On the other hand, the relative share on the total information of a 
single feature decreases with higher $p_\text{rel}$. For small $\beta$, the 
distribution of $\Delta\text{RMSE}_\text{rel}$ is very skewed and the first 
effect dominates. For larger $\beta$, the distribution becomes less skewed and 
the latter effect has a higher impact. In total, this results in the observed 
trends with increasing $p_\text{rel}$.

Altogether, our paper addressed implications of using the benefit-cost ratio
without an additional hyperparameter for cost-sensitive feature selection. As 
using this ratio is a typical approach to incorporate feature costs, it is 
important to understand possible problems resulting from it. We provided a 
thorough problem description and analyzed multiple parameter settings in a 
simulation study. Results from this study illustrated that a strong 
cost-scaling, which may result from high relative cost differences between 
features, can notably influence the detection limit of relevant features. This 
effect interacts with the number of noise features in the data.

To avoid this problem we recommend using an adapted benefit-cost ratio, such as 
the ones proposed in \citet{jagdhuber2020cost} or \citet{min2011test}. 
The main alternative solution to incorporate costs is a weighted linear 
combination as mentioned in the introduction of this paper. All these 
approaches share the idea of introducing a hyperparameter to control the 
trade-off between benefit and cost. This can reduce the problem, but it comes 
at the price of an additional estimation step. If the analysis requires the 
benefit-cost ratio without hyperparameter, we strongly recommend to thoroughly 
analyze the cost distribution of the given data set. If relative cost differences 
are high, transforming costs prior to applying the benefit-cost ratio may be 
beneficial. In practice, such extreme ratios may likely occur with some costs 
very close to 0, or from setting cost-free features to a cost of $\epsilon$ 
close to 0, as e.g. recommended in \citet{min2011test}.

The popularity of the benefit-cost-ratio shows the need for simple methods to
incorporate costs without an additional parameter tuning step. Beyond the scope
of this work, solving this problem with a comprehensible way to specify the
trade-off between costs and performance with expert knowlegde, instead of 
tuning a black-box hyperparameter, would be of great interest. This would allow 
the user to specify the intended relation of costs and performance, which may 
differ greatly between fields of application. Our work covers a specific task 
in predictive modelling and tries to raise awareness of the problem. Further 
research may also consider different model types, performance measures, feature 
distributions, and additional aspects.

\section*{Appendix}
\subsection*{Additional file 1}\label{add:1}
\textit{Extended version of Figure~\ref{fig:selprob}. (PDF)}\\
Selection probabilities of relevant, noise or no feature, along multiple values 
of the cost-scaling factor $\theta$ (columns), the number of noise features 
$p_\text{noise}$ (rows), and the effect size of relevant features $\beta$ 
(x-axis). Pages describe different settings of $p_\text{rel} = 1, 2, 5, 10$.

\subsection*{Additional file 2}\label{add:2} 
\textit{Extended version of Figure~\ref{fig:empdist}. (PDF)}\\
Empirical distributions of $\Delta\text{RMSE}$ for noise features (red) and
relevant features (blue). The latter are illustrated as heatmap along different 
values of the true effect size $\beta$. Lighter colors indicate higher 
densities. Each page describes a combined setting of 
$p_\text{rel} = 1, 2, 5, 10$ and $\theta = 1, 10, 100, 1000$. Contrary to 
Figure~\ref{fig:empdist}, plots are rescaled to illustrate the empirical 
distribution of the relevant features.



\section*{Declarations}
\subsection*{Funding}
This work was supported by Deutsche Forschungsgemeinschaft (DFG), Project
RA 870/7-1, and Collaborative Research Center SFB 876, A3. The authors
acknowledge financial support by Deutsche Forschungsgemeinschaft and
Technische Universität Dortmund within the funding programme Open
Access Publishing. 

\subsection*{Conflict of interest}
The authors declare that they have no conflict of interest.

\subsection*{Availability of data and material}
The datasets used and/or analysed during the current study are available 
from the corresponding author on reasonable request.

\subsection*{Code availability}
The full code used during the current study is available from the 
corresponding author on reasonable request.

\subsection*{Authors contributions}
Rudolf Jagdhuber initiated the topic, formulated and discussed the problem, 
designed and executed the simulation studies, interpreted the results, and 
wrote the manuscript.
Jörg Rahnenführer supervised the project, contributed to the problem 
definition, the design of the simulation study and to the interpretation of 
the results, and corrected and approved the manuscript.

\bibliographystyle{plainnat}

\bibliography{Manuscript_wFigures}



\end{document}